\documentclass[10pt,twocolumn,letterpaper]{article}

\usepackage{cvpr}              %

\usepackage{graphicx}
\usepackage{amsmath}
\usepackage{amssymb}
\usepackage{booktabs}
\usepackage{enumitem}
\usepackage{microtype}

\usepackage[pagebackref,breaklinks,colorlinks]{hyperref}

\newcommand{\footremember}[2]{%
   \thanks{\xspace\xspace#2}
    \newcounter{#1}
    \setcounter{#1}{\value{footnote}}%
}

\usepackage{amssymb}
\usepackage{amsmath,amsfonts}
\usepackage{amsopn,bm,mathtools}

\usepackage{nicefrac}       %

\newcommand{\ProbOpr}[1]{\mathbb{#1}}

\newcommand{\expect}[2]{%
\ifthenelse{\equal{#2}{}}{\ProbOpr{E}_{#1}}
{\ifthenelse{\equal{#1}{}}{\ProbOpr{E}\left[#2\right]}{\ProbOpr{E}_{#1}\left[#2\right]}}} %
\newcommand{\var}[2]{%
\ifthenelse{\equal{#2}{}}{\ProbOpr{VAR}_{#1}}
{\ifthenelse{\equal{#1}{}}{\ProbOpr{VAR}\left[#2\right]}{\ProbOpr{VAR}_{#1}\left[#2\right]}}} %

\DeclareMathOperator{\argmin}{arg\,min}

\usepackage{xspace}

\makeatletter
\DeclareRobustCommand\onedot{\futurelet\@let@token\@onedot}
\def\@onedot{\ifx\@let@token.\else.\null\fi\xspace}
\def\eg{\emph{e.g}\onedot} 
\def\ie{\emph{i.e}\onedot}

\makeatother

\usepackage{mathtools}

\usepackage[ruled,vlined]{algorithm2e}

\usepackage{listings}

\graphicspath{{../figs/}{../}}

\newcommand{\eat}[1]{}

\usepackage{makecell}

\newcommand{\ourtitle}{{Co-training Transformer with Videos and Images Improves Action Recognition}}

\newcommand{\ourmethod}{{\textsc{CoVeR}}\xspace}

\usepackage{xcolor}
\definecolor{codegreen}{rgb}{0,0.6,0}
\definecolor{codegray}{rgb}{0.5,0.5,0.5}
\definecolor{codepurple}{rgb}{0.58,0,0.82}
\definecolor{backcolour}{rgb}{0.95,0.95,0.92}
\definecolor{darkgreen}{rgb}{0,0.4,0}
\definecolor{cerise}{rgb}{0.871, 0.192, 0.388}
\definecolor{carmine}{rgb}{0.59, 0.0, 0.09}
\definecolor{olive}{rgb}{0.332, 0.418, 0.184}
\definecolor{navyblue}{rgb}{0.496, 0.810, 0.837}

\definecolor{darkergreen}{RGB}{21, 152, 56}
\definecolor{red2}{RGB}{252, 54, 65}

\usepackage{amssymb}%
\usepackage{pifont}%

\usepackage{multirow}

\makeatletter
\@namedef{ver@everyshi.sty}{}
\makeatother
\usepackage{tikz}

\usepackage[colorinlistoftodos,linecolor=blue,backgroundcolor=white,bordercolor=blue]
{todonotes}

\begin{document}

\title{\ourtitle}

\author{
  Bowen Zhang \footremember{Google}{This work was conducted at Google Brain.} \\
  USC \\
  \small\texttt{zhan734@usc.edu} \\
  \and
  Jiahui Yu \\
  Google Brain\\
  \small\texttt{jiahuiyu@google.com} \\
  \and
  Christopher Fifty \\
  Google Brain \\
  \small\texttt{cfifty@google.com} \\
  \and
  Wei Han \\
  Google Brain \\
  \small\texttt{weihan@google.com} \\
  \and
  Andrew M. Dai \\
  Google Brain \\
  \small\texttt{adai@google.com} \\
  \and
  Ruoming Pang \footremember{Apple AI}{Part of this work was conducted while at Google Brain} \\
  Apple AI\\
  \small\texttt{ruoming@gmail.com} \\
  \and
  Fei Sha \\
  Google Research\\
  \small\texttt{fsha@google.com} \\
}

\maketitle

\begin{abstract}
\vspace{-10pt}
In learning action recognition, models are typically pre-trained on object recognition with images, such as ImageNet, and later fine-tuned on target action recognition with videos. This approach has achieved good empirical performance especially with recent transformer-based video architectures. While recently many works aim to design more advanced transformer architectures for action recognition, less effort has been made on how to train video transformers. In this work, we explore several training paradigms and present two findings. First, video transformers benefit from joint training on diverse video datasets and label spaces (\eg, Kinetics is appearance-focused while SomethingSomething is motion-focused). Second, by further co-training with images (as single-frame videos), the video transformers learn even better video representations. We term this approach as Co-training Videos and Images for Action Recognition (\ourmethod). In particular, when pretrained on ImageNet-21K based on the TimeSFormer architecture, \ourmethod improves Kinetics-400 Top-1 Accuracy by 2.4\%, Kinetics-600 by 2.3\%, and SomethingSomething-v2 by 2.3\%. When pretrained on larger-scale image datasets following previous state-of-the-art, \ourmethod achieves best results on Kinetics-400 (87.2\%), Kinetics-600 (87.9\%), Kinetics-700 (79.8\%), SomethingSomething-v2 (70.9\%), and Moments-in-Time (46.1\%), with a simple spatio-temporal video transformer.

\end{abstract}

\section{Introduction}
\label{sec:intro}
Action recognition has received significant attention from the research community, as many applications can benefit from improved action recognition modeling, such as video retrieval~\cite{ging2020coot,krishna2017dense}, video captioning~\cite{krishna2017dense}, video QA~\cite{li2021value}, etc.
Datasets are one dimension of improvement. Video datasets have evolved from hundreds of videos in controlled environments~\cite{schuldt2004recognizing} to millions of videos crawled from the Internet~\cite{monfort2019moments,kay2017kinetics}. In addition to quantity, the scope of videos has also broadened. For example, the topics covered by action recognition datasets have evolved from simple body motions like waving and handshaking to more complex activities present in our daily life. Simultaneously, with the increase of data and class distribution complexity, modeling architecture complexity has likewise increased~\cite{simonyan2014two,wang2021tdn,carreira2017quo,bertasius2021space,arnab2021vivit,fan2021multiscale,liu2021video}. Among these architectures, Transformer based approaches have recently demonstrated state-of-the-art performance on several benchmarks~\cite{liu2021video,bertasius2021space,zhang2021vidtr,arnab2021vivit}. However, since Transformer models are more data-hungry~\cite{touvron2021training} and action recognition datasets are relatively small in scale, large Transformer models are typically first trained on image datasets and later finetuned on the target action recognition dataset. 

\begin{table}[]
\centering
\begin{tabular}{@{}l@{\quad}ccc@{}}
\toprule
Training Policy & Standard & \ourmethod & $\Delta$  \\
\midrule
Kinetics-400    & 80.7 & \textbf{83.1} & \textbf{\textcolor{darkergreen}{+2.4}}\\
Kinetics-600    & 82.2 & \textbf{84.5} & \textbf{\textcolor{darkergreen}{+2.3}}\\
Kinetics-700    & - & \textbf{74.9} & - \\
SSv2            & 62.4 & \textbf{64.7} & \textbf{\textcolor{darkergreen}{+2.3}}\\
Moments-in-Time & - & \textbf{41.5} & - \\
\bottomrule

\end{tabular}
\caption{Top-1 accuracy comparison between standard training policy and \ourmethod using TimeSFormer pretrained on ImageNet-21k (I21K). \ourmethod co-trained with multiple image and video datasets achieves better performance on all datasets.}
\label{tab:teaser}
\vspace{-10pt}
\end{table}

While the current pre-training and fine-tuning action recognition paradigm is straightforward and manifests strong empirical results, it may be overly restrictive for building general-purpose action-recognition models. Compared to a dataset like ImageNet~\cite{deng2009imagenet} that covers a large range of object recognition classes, action recognition datasets like Kinetics~\cite{kay2017kinetics} and Something-Something-v2 (SSv2)~\cite{goyal2017something} pertain to limited topics. For example, Kinetics focuses on actions like ``cliff diving'' and ``ice climbing'' while SSv2 contains information related to object agnostic activities like ``pretending to put something onto something else.'' As a result, adapting an action recognition that has been fine-tuned on Something-Something-v2 to a disparate dataset like Kinetics is likely to result in poor performance. Differences in objects and video backgrounds among datasets further exacerbate learning a general-purpose action recognition classification model, and even though video datasets may be increasing in size, prior work~\cite{arnab2021vivit,fan2021multiscale,liu2021video} suggests significant data augmentation and regularization is necessary to achieve strong performance. This latter finding may indicate the model quickly overfits on the target dataset, and as a result, hinders its capacity to generalize to other action recognition tasks.

In this work, we aim to build a training strategy for a general purpose action recognition model. Inspired by prior works in vision and language that demonstrate a single Transformer model can be extended to many downstream tasks~\cite{su2019vl,lu2019vilbert}, we propose to leverage both image and video data to jointly train a single action recognition model. This approach is buttressed by two main findings. First, disparate video datasets cover a diverse set of activities, and training them together in a single model could lead to a model that excels in a wide range of activities. 
Second, video is a perfect source for learning motion information, while images are great for exploiting  appearance structure. Compared to ImageNet, action recognition datasets have relatively little data. Combining this with the high spatial redundancy among frames in a clip, and leveraging a diverse distribution of image examples may be beneficial in building robust spatial representations in video models.

With this background as motivation, we suggest a new training scheme: co-training video and image for action recognition (\ourmethod). Similar to the typical pre-training and fine-tuning paradigm, \ourmethod first pre-trains the model on an image dataset, but during fine-tuning, it simultaneously trains a single model on multiple action recognition and image datasets to build robust spatial and temporal representations of video data. More concretely, \ourmethod adopts a multi-task learning setup to train multiple datasets within one model, with the total loss for a given batch equal to the weighted sum of the losses across each dataset in that batch. Our empirical findings suggest this approach is competitive across several action recognition benchmarks. Moreover, unlike the current paradigm of pre-train once and fine-tune on each downstream action recognition benchmark, \ourmethod learns generalizable representations which can be used on downstream action recognition tasks without any additional finetuning. Our empirical findings in Table~\ref{tab:teaser} indicate this simplified approach leads to improved performance on Kinetics, SomethingSomething-v2, and Moments-in-Time when compared with the typical pre-train and fine-tune paradigm.

 \textbf{Our contributions} are three-fold: 
 \begin{itemize}[leftmargin=*,topsep=1pt,itemsep=0pt]
     \item We analyze simultaneous training of image and video data for the purpose of modeling action recognition data. 
     \item We propose \ourmethod, an approach that learns robust spatial and temporal representations via simultaneous learning across multiple action recognition and image datasets. The learned representations can be immediately applied to several downstream tasks to perform competitively with prior pre-training and fine-tuning paradigms. 
     \item \ourmethod  established new State-of-The-Art results on multiple datasets.
 \end{itemize}

\section{Related Work}
\label{sec:related}

\begin{figure*}[t]
\begin{subfigure}{.45\textwidth}
  \centering
  \includegraphics[width=.73\linewidth]{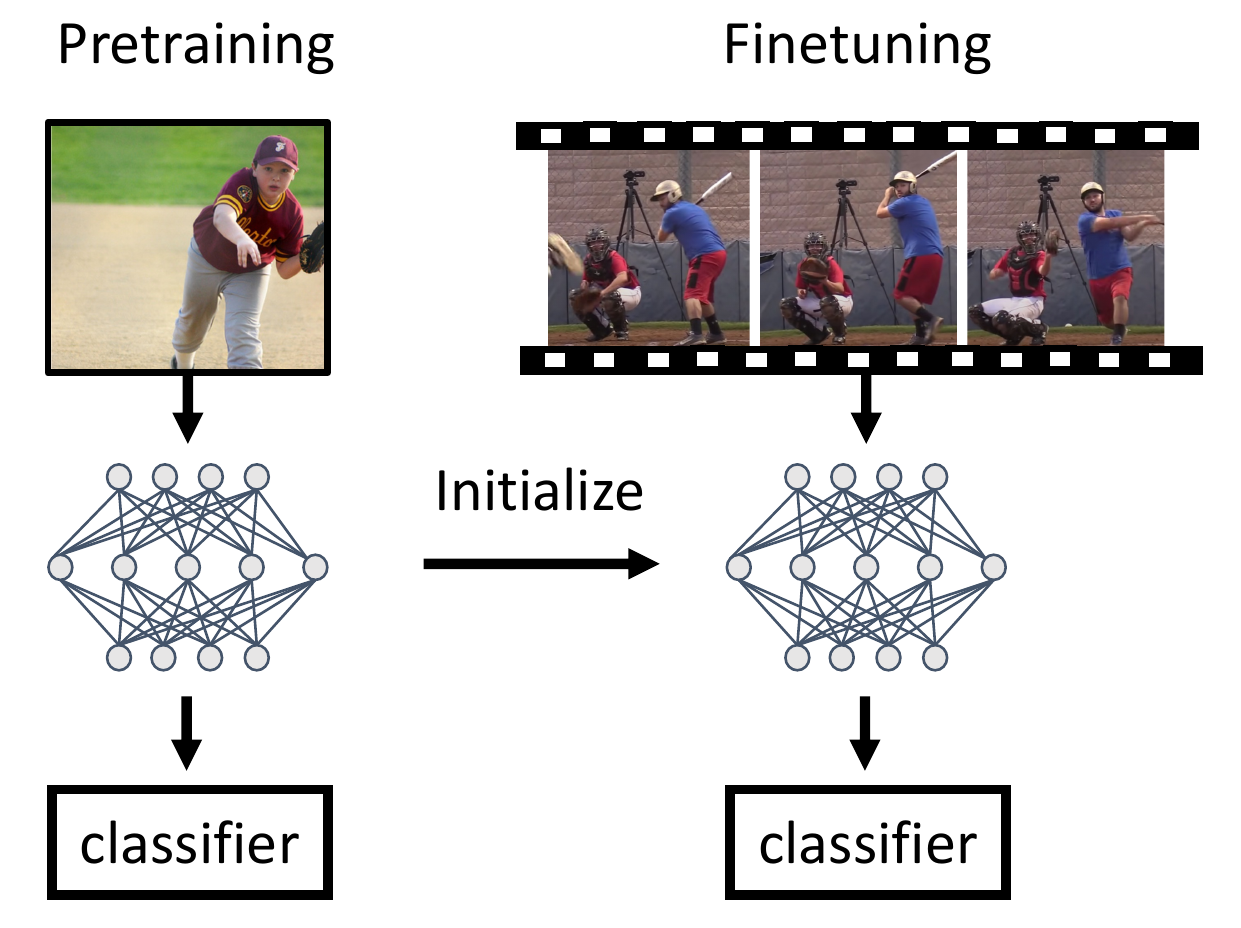}  
  \caption{Standard training paradigm}
\end{subfigure}
\begin{subfigure}{.55\textwidth}
  \centering
  \includegraphics[width=\linewidth]{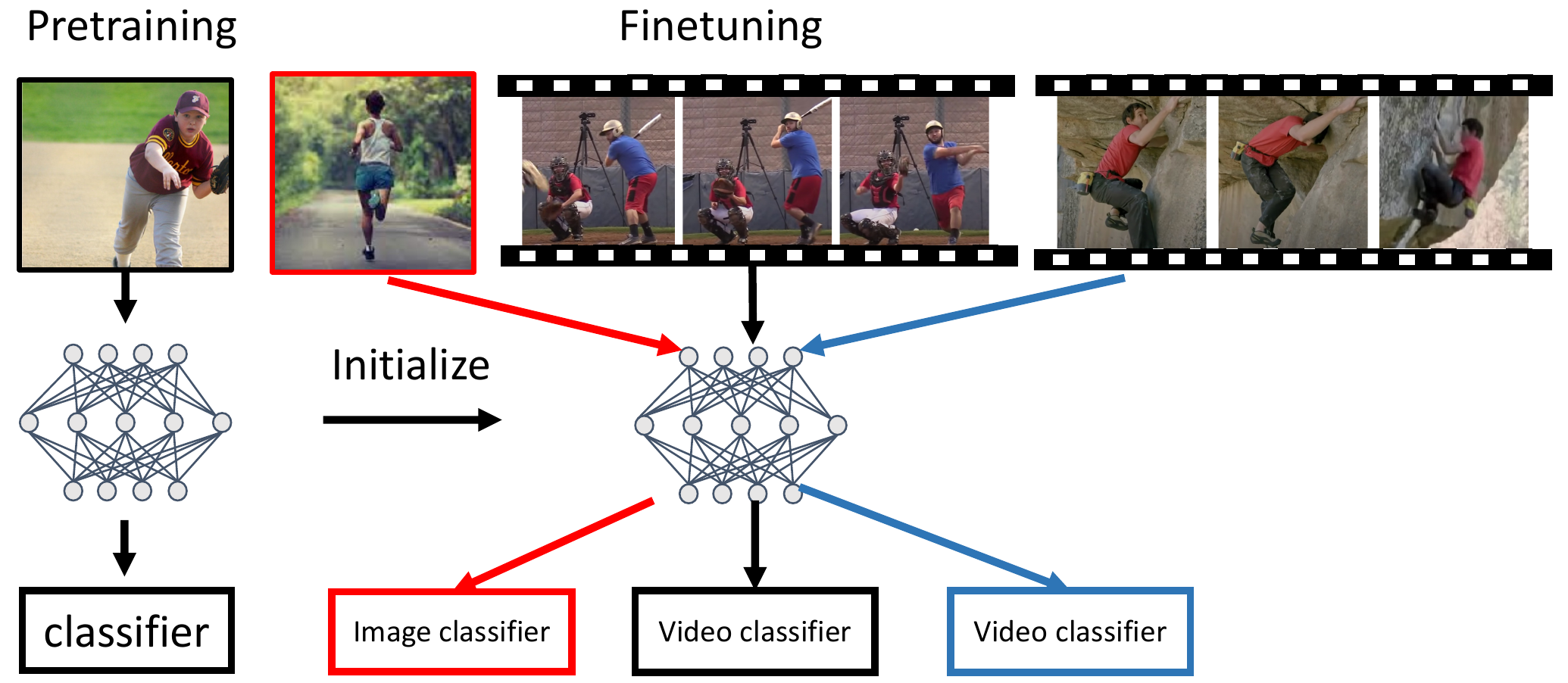}  
  \caption{\ourmethod training paradigm}
\end{subfigure}
\caption{The comparison between the proposed \ourmethod and the standard training paradigm. The difference is \ourmethod co-trained on multiple image and video datasets during finetuning, while the standard training paradigm only finetuned on one dataset.}
\label{fig:cover}
\vspace{-10pt}
\end{figure*}

Action recognition modeling persists as a challenging problem in the field of computer vision. Recent work in this domain largely focus on two dimensions to improve performance: modeling changes and training paradigm. 

\paragraph{Video Transformers} 
Recently, and inspired by the Visual Transformer~\cite{dosovitskiy2020image} and BERT\cite{devlin2018bert} in the domain of Natural Language Processing, action recognition modeling has begun to adopt transformer-based architectures such as TimeSFormer~\cite{bertasius2021space}, ViViT~\cite{arnab2021vivit}, and Multiscale Vision Transformer~\cite{fan2021multiscale}. In this class of models, TimeSFormer and ViViT directly extend the Visual Transformer into the video domain by adding temporal attention layers. Based on this architecture, TimeSFormer and ViViT find that large-scale image data pretraining is beneficial, and with this training policy, they can surpass the performance of ConvNet-based architectures \cite{carreira2017quo,simonyan2014two,tran2015learning}. Further exploring training policies, Multiscale Vision Transformer finds that, with carefully tuned parameters, strong performance can be achieved without pre-training on a large-scale image dataset. A separate approach, and inspired by the Swin Transformer~\cite{liu2021swin}, Video Swin-Transformer~\cite{liu2021video} adopts 3D shifted windows to learn spatial and temporal aware video representations. While modeling approaches have undoubtedly driven much of the recent advancement of action recognition, we instead direct our focus to these models' training policy and how training on varying data distributions may learn general-purpose action recognition models.

\paragraph{Training Paradigms} We define training policy as the technique(s) used to train a model. In this sense, and for the domain of action recognition, 
Two-Stream ConvNet~\cite{simonyan2014two} and I3D~\cite{carreira2017quo}
were the first approaches to leverage image data to improve video modeling. More concretely, both models leveraged the feature extractors learned on image data during ``pre-training'' and would later ``fine-tune'' on a downstream video benchmark. The concept of using image data to learn spatial relationships which can be transferred to video understanding has also been applied to recent transformer-based approaches \cite{bertasius2021space,arnab2021vivit,liu2021video}. For instance, OmniSource~\cite{duan2020omni} extends 3D ConvNet by pre-training on both image and video datasets, and then later fine-tuning on a target dataset. 
Similarly, UniDual is also tailored to ConvNets and proposes fine-tuning on one image and one video dataset to improve video modeling performance. Unlike OmniSource and UniDual, \ourmethod is tailored to transformer-based architectures. More specifically, OmniSource and UniDual require an additional layer to process image data, whereas \ourmethod directly processes video and image data without any modification to the model architecture. Moreover, OmniSource fine-tunes the learned weights on each down-stream dataset during evaluation. Our approach simplifies this process by building representations that are generalizable across multiple datasets. UniDual fine-tunes the model on a single image and a single video dataset. This differs from \ourmethod that explores learning multiple video datasets simultaneously. Finally, our empirical results suggest \ourmethod outperforms OmniSource and UniDual on several action recognition datasets.

\vspace{-5pt}
\section{Method}
\vspace{-5pt}
\label{sec:method}

To motivate our suggested change, we first describe the typical spatio-temporal transformer framework, paying special attention to improving generalisation by pre-training on image data and fine-tuning on video data. Next, we build on this analysis to present co-training image and video for action recognition (\ourmethod). \ourmethod changes the typical training paradigm by leveraging multiple image and video datasets to train a spatio-temporal transformer model.

\subsection{Video Transformer for Action Recognition}
The action recognition framework is composed of two components, an action recognition model $f$ and a training policy. In this paper, we describe a simple spatio-temporal attention factorized transformer (\ie, TimeSFormer\cite{bertasius2021space}) as a prototypical action recognition model.

\vspace{-10pt}
\paragraph{TimeSFormer}
TimeSFormer is an extension of Visual Transformer\cite{dosovitskiy2020image}. Similar to a ViT model, TimeSformer can be reduced to a sequence of self-attention blocks; however rather than using a single spatial self-attention mechanism, TimeSformer augments the self-attention block with a temporal attention mechanism. Analogous to the Transformer model introduced for natural language processing, both spatial and temporal attention mechanisms are formulated as Multi-Head Attention (MHA). MHA takes three sets of input elements, \ie, the key set $K$, the query set $Q$, and the value set $V$, and performs scaled dot-product attention as:

\vspace{-10pt}
\begin{equation}
    \texttt{MHA}(K, Q, V) = \texttt{FFN}\big(\texttt{Softmax}(\frac{Q^\top K}{\sqrt{d}}) \cdot V \big) \nonumber
\end{equation}
Here, \texttt{FFN} is a feed-forward neural network and $d$ is the dimension of $K$ and $Q$. With different choices of $K$ and $V$, MHA can be categorized as either spatial attention or temporal attention. Spatial attention corresponds to $K$ and $V$ sampled from the same frame and temporal attention corresponds to $K$ and $V$ sampled across the frames of a video clip. Each TimeSFormer block contain one layer of temporal attention and one layer of spatial attention.

TimeSFormer takes $n$ video frames $x_\texttt{video}$ as input. The frames are first uniformly cropped into a sequence of image patches with size of $(n, s, h, w)$, where $s$ is the number of image patches within one frame. $h$ and $w$ represent the spatial resolution. The image patches are then fed into $L$ TimeSFormer blocks through multiple spatial attention and temporal attention layers. An affine transformation is then applied to the learned representation $t_\texttt{video}=f(x_\texttt{video})$ to attain a probability distribution across all label classes $c_\texttt{video} = \texttt{MLP}(t_\texttt{video})$.

In this paper, we describe TimeSFormer using $\theta_s$ and $\theta_t$, where $\theta_s$ represents the parameters within spatial attention layers and $\theta_t$ represents the parameters within the temporal attention layers. We describe the classification layer as $\theta_{\texttt{MLP}}$.

\vspace{-10pt}
\paragraph{Standard Training Paradigm}
Due to the large amount of parameters and the limited size of the video dataset, the standard training policy follows a classical pretraining and finetuning approach. The model is first pretrained on a large object recognition image dataset $D_{\texttt{image}}$ and then finetuned on the target downstream video dataset $D_{\texttt{video}}$. Specifically, during the pre-training stage, the temporal attention layers are all removed. Only the parameters of the spatial attention layers $\theta_s$ are optimized, by minimizing the training loss 
\begin{equation}
    \theta_s = \argmin_{\theta_s} \ell (\{y_{\texttt{image}}\}, \{c_\texttt{image}\})
    \label{image_cls}
\end{equation} where  $\{c_\texttt{image}\} = \texttt{MLP}(f(\{x_\texttt{image}\}; \theta_s))$ is the classification probability. $\{x_{\texttt{image}}\}$ and $\{y_{\texttt{image}}\}$ denote the videos and labels in a mini-batch, and are randomly sampled from the image dataset $D_{\texttt{image}}$. $\ell$ is the cross-entropy loss function.

After pre-training, both spatial attention layers and temporal attention layers are finetuned on the target video datasets.
\begin{equation}
    (\theta_s, \theta_t) = \argmin_{\theta_s, \theta_t} \ell (\{y_{\texttt{video}}\}, \{c_\texttt{video}\}) \nonumber
\end{equation} where $(x_{\texttt{video}}, y_{\texttt{video}})$ are sampled from the video dataset $D_{\texttt{video}}$.

\subsection{Action Recognition Analysis}

A robust learned representation of video data should be descriptive in both the spatial and temporal dimension. Our empirical findings suggest the typical pre-training and fine-tuning paradigm may limit the model's capacity to construct generalizable representations by fine-tuning on a single, and relatively small, downstream action recognition task.

Expanding on this last point, it is likely each video dataset incorporates a dataset-specific bias. Applying this hypothesis to two popular video datasets, K400 and SSv2, we find SSv2 focuses on object interaction and relies on complex temporal reasoning to achieve strong performance~\cite{arnab2021vivit}. On the other hand, K400 focuses on interactions among humans and objects. Given the strong performance of non-temporal action recognition models on this dataset, complex temporal reasoning may be significantly less important than learning robust representations comprised of spatial information.

\begin{table}[t]
\centering
\begin{tabular}{@{}lccr@{}}

\toprule
Frame order & Normal  & Reversed & $\Delta$\\
\midrule
K400 & \textbf{78.1} & 78.0 & \textbf{\textcolor{red}{-0.1}} \\
SSv2 & \textbf{58.8} & 22.6 & \textbf{\textcolor{red}{-36.2}}\\
\bottomrule
\end{tabular}
\caption{Performance of TimeSFormer trained using $224\times224$ image resolution and evaluated on normal frame order and reversed frame order. Reversed frame order means the order of frames are reversed during test-time.}
\label{tab:shuffle_test}
\vspace{-10pt}
\end{table}
To further analyze the inherent dataset bias hypothesis, we reverse the order of frames in a clip during test-time. We follow the standard paradigm by training a TimeSformer model~\cite{bertasius2021space} using a 224$\times$224 image resolution. We report our findings with both the normal and reversed ordering of frames along the temporal dimension in Table \ref{tab:shuffle_test}. Our findings indicate the model may learn strong temporal-based representations when training on SSv2, while temporal information appears less important for K400. For example, the difference in test accuracy on K400 is small at -0.1\%, but the difference in test accuracy on SSv2 is -36.2\%.

Another facet of our analysis relates to the limited size and scope of publicly available video datasets. Given the difference in focus of each dataset, the representations learned on one dataset distribution may not generalize to that of other datasets. To analyze this dimension, we conduct an experiment by training a TimeSformer model to achieve strong performance on the Kinetics-400 dataset and then reinitialize the classification layer, freeze the feature extraction layers, and fine-tune the model on SSv2. Our findings indicate this approach would only yield an accuracy of 19.5\% on SSv2, and we interpret this result to indicate the feature extraction mechanisms learned by one action recognition model may have difficulty generalizing to other datasets.

A possible solution to mitigate inherent dataset biases would be to collect a single large-scale video dataset covering a diverse range of actions and events. However, such a collection would be challenging to design and time-consuming to create. An added layer of complexity relates to deciding the label classes of this dataset, and mapping a single video to one or more classes is non-trivial and would require careful design. A different approach is to learn representations applicable to many disparate action recognition datasets. Rather than learning video representations specific to a single dataset during fine-tuning, we instead suggest learning a single model across many action recognition dataset distributions.

\subsection{\ourmethod: Co-train Videos and Images for Action Recognition}

\ourmethod leverages different action recognition dataset distributions, as well as large-scale object recognition, to construct a general-purpose feature-extraction model. We first introduce the mechanism of learning representations suited for multiple video datasets and then describe the process of integrating image data into the fine-tuning paradigm.

\vspace{-10pt}

\subsubsection{Co-train Videos}

\label{sec:defacto}
\begin{figure*}[t]
  \centering
  \includegraphics[width=\linewidth]{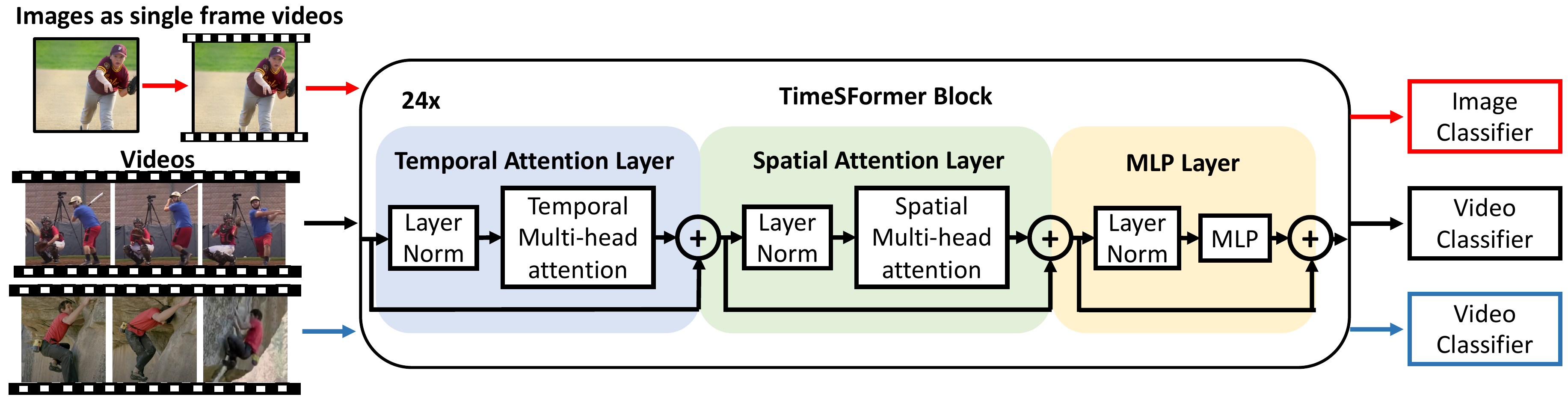}  
\caption{\ourmethod adopts multi-task learning strategy. Each dataset has its own classifier. For the image dataset, we consider images as single frame videos. Therefore, the temporal multi-head attention will not affect the image input.}
\label{fig:cover}
\vspace{-12pt}
\end{figure*}

To learn from $N$ video datasets, we adopt a multi-task learning paradigm and equip the action recognition model $f$ with $N$ classification heads $\{\texttt{MLP}_i\}_{i=1}^n$. However, pre-training is unchanged. Notably, we adopt the typical pre-training policy of learning all non-temporal parameters on a large-scale object recognition dataset by minimizing the coss entropy loss over Eq. \ref{image_cls}. 

In the co-training policy, we follow the pretraining and finetuning strategy. Similar to the standard training policy, the spatial attention layers $\theta_s$ is learnt by minimizing the cross entropy loss over Eq. \ref{image_cls}.

During fine-tuning, \ourmethod learns both spatial and temporal attention layers across the samples $(x_{\texttt{video}}^i, y_{\texttt{video}}^i) \sim D_{\texttt{video}}^i$ from $N$ datasets jointly, where $(x_{\texttt{video}}^i, y_{\texttt{video}}^i)$ is the video and its label sampled from the dataset $i$. All video samples are processed by the model $f$ via the shared parameters $\theta_s$ and $\theta_t$. The sample representations are then distributed to the appropriate classification head to obtain the classification probability $c_\texttt{video}^i = \texttt{MLP}(f(x_\texttt{video}^i; \theta_s, \theta_t))$. We calculate the training loss for samples in video dataset $i$ as

\begin{equation}
    \ell_\texttt{video}^i = \ell (\{y_{\texttt{video}}^i\}, \{c_\texttt{video}^i\}) \nonumber
\end{equation}

To optimize the parameters $\theta_s$ and $\theta_t$, we minimize the weighted sum of the loss function across all $N$ datasets.
\begin{equation}
    (\theta_s, \theta_t) = \argmin_{\theta_s, \theta_t} \sum_i w^i \cdot \ell_\texttt{video}^i \nonumber
\end{equation}
where $w^i$ is the weight for the loss function of the dataset $i$.

Jointly learning action recognition feature extractors across multiple video datasets conveys two advantages. First, as the model is directly trained on multiple datasets, the learned video representations are more general and can be directly evaluated on those datasets without additional fine-tuning. Second, and as emphasized in prior work~\cite{bertasius2021space}, there may be benefits from expanding the scope and quantity of action recognition examples. Attention-based models may easily overfit to a smaller video distribution, thus degrading the generalization of the learned representations. Training on multiple datasets mitigates this challenge by reducing the risk of overfitting. Finally, as indicated in Sect. \ref{sec:defacto}, certain datasets may focus on different inductive biases of video modeling. For example, one dataset may emphasize the modeling of temporal information while others emphasize spatial representational learning. Jointly learning on both distributions may lead to more robust feature extractors that encode both appearance and motion information to improve performance on action recognition benchmarks.

\subsubsection{Co-train Video and Image Data}

To maintain strong spatial representations, \ourmethod trains a model $f$ on both image and video datasets. Similar to the training policy of transformer-based video models, we first pre-train spatial attention layers $\theta_s$ using a large object recognition dataset, and then fine-tune the entire model $(\theta_s, \theta_t)$ using both video datasets $D_\texttt{video}^i$ and image datasets $D_\texttt{image}^j$ as Fig.~\ref{fig:cover}.

We adapt an object recognition image task to an action recognition video paradigm with minimal modification by considering an image as a video with only one frame. In this context, we can directly create a batch of both image $x_\texttt{image}^j\sim D_\texttt{image}^j$ and video $x_\texttt{video}^i\sim D_\texttt{video}^i$ data as input to the TimeSformer model $f$. With regards to the object recognition task, we obtain classification outputs $c_\texttt{image}^j =  \texttt{MLP}(f(x_\texttt{image}^j; \theta_s, \theta_t))$, and for the video datasets, we denote dataset-specific classification outputs as $c_\texttt{video}^j =  \texttt{MLP}(f(x_\texttt{video}^j; \theta_s, \theta_t))$. The weighted loss for co-training over both images and videos is

\begin{align}
    \ell_\texttt{image\_video} =& \sum_i w_\texttt{video}^i \cdot \ell (\{y_{\texttt{video}}^i\}, \{c_\texttt{video}^i\}) \nonumber \\
    &+ \sum_j w_\texttt{image}^j \cdot \ell (\{y_{\texttt{image}}^j\}, \{c_\texttt{image}^j\}) \nonumber
\end{align}
where $w_\texttt{image}^j$ and $w_\texttt{video}^i$ represents the loss weights for the image dataset $j$ and the video dataset $i$. We minimize  $\ell_\texttt{image\_video}$ to optimize parameters $\theta_s$ and $\theta_t$.

The comparison between the standard training policy and \ourmethod are summarized in Fig.~\ref{image_cls}

\subsection{Discussion}
An ideal video representation should capture both the appearance and motion information from a video. Although video datasets are informational sources for learning motion information, the spatial information contained within a video clip may be limited. This is due to redundancy among frames and the relatively ``small'' size of video datasets compared to classical image datasets. Therefore, even by simultaneously training a model on many video datasets, the model's capacity to learn appearance information may be hindered. Although image dataset pretraining may provide a good start for obtaining the appearance knowledge, it is possible the robust spatial representations are diluted during fine-tuning on highly spatially redundant video data. Reducing the robustness of learned spatial representations would likely decrease model performance; however, these representations may be maintained by continuing to train object recognition with action recognition during the fine-tuning stage of learning.

\section{Experiment}
\label{sec:exp}

In this section, we first present the experiment setup and the implementation details. Then we study the performance of \ourmethod on five large-scale video datasets. 

\subsection{Experiment Setup}
\label{subsec:setup}

\paragraph{Datasets.}We evaluate our approach on 5 challenging video datasets, Kinetics-400 (K400)~\cite{kay2017kinetics}, Kinetics-600 (K600)~\cite{carreira2018short}, Kinetics-700 (K700)~\cite{carreira2019short}, SomethingSomething-v2 (SSv2)~\cite{goyal2017something}, and Moments in Time (MiT)~\cite{monfort2019moments}. The Kinetics dataset is a collection of datasets containing three variants, Kinetics-400, Kinetics-600, and Kinetics-700 with 400, 600, and 700 classes respectively. The Kinetics dataset focuses on daily human-object interaction. The SSv2 dataset contains 174 classes with 168K videos for training and 24K videos for evaluation. It contains videos with object-agnostic action. The Moments-in-Time dataset is one of the largest datasets for video understanding. The dataset contains 791K videos for training and 33K videos for evaluation, which cover human actions as well as animal actions. The Moments-in-Time dataset covers a wide range of videos from Youtube to cartoons. For all datasets, we use the standard training and testing splits. We report the results on the testing split for all datasets.

\vspace{-10pt}
\paragraph{Implementation details.} 

We conducted experiments using TimeSFormer~\cite{bertasius2021space}, as it achieves the balance between efficiency and performance.  The TimeSFormer model contains 24 TimeSFormer blocks. Within each block, there is one temporal multi-head attention layer followed by one spatial multi-head attention layer. Each multi-head attention layer has 16 attention heads with 1024 hidden dimensions in total. Note that our co-training policy could also be applied to other transformer-based approaches~\cite{liu2021video,arnab2021vivit} without further modification. 

Given a video, we first uniformly sampled 16 frames across the entire video. Then image patches with the resolution of 448$\times$448 are randomly cropped from the sampled frames to form the input. Similarly, when co-training with images, we randomly cropped a 448$\times$448 image patch. We only applied random cropping and random horizontal flipping to augment the training data during training. Note that the horizontal flipping is not applied to SSv2 videos, as two of the SSv2 action categories, ``Pushing something from left to right'' and ``Pushing something from right to left'', are symmetric. Following TimeSFormer, we evaluate our model by averaging the predictions from 1$\times$3 views of cropping. Specifically, during evaluation, we obtain three spatial crops from video. We didn't do temporal cropping during the evaluation. 

We study the performance under three pre-training image datasets, ImageNet-21k~\cite{deng2009imagenet}, JFT-300M~\cite{kolesnikov2020big}, and JFT-3B~\cite{zhai2021scaling}. We finetune the pre-trained model on video and image datasets using a mini-batch of 128. Within each batch, we sampled videos and images from all finetuning datasets. The sampling rate is proportional to the size of the datasets. The model is optimized using SGD with momentum set to 0.9. The model is trained for 20 epochs. The initial learning rate is set as 5$e-$3. It drops to 5$e-$4 and 5$e-$5 at epochs 11 and 15, respectively. 

\subsection{Main results}

\begin{table*}[t]
    \centering

\begin{tabular}{@{}ll@{\quad\quad}cl@{~}l@{~}l@{\quad~}l@{~}l@{\quad\quad}c@{}}
\toprule
Model                 & Pretrain & Finetune        & K400 & K600 & K700 & SSv2 & MiT & Views \\
\midrule                      
Video SwinTrans~\cite{liu2021video}      & ImageNet21k & K400  & \textbf{84.9}              &      \multicolumn{1}{l}{-}    & \multicolumn{1}{l}{-}       &    \multicolumn{1}{l}{-}    &      \multicolumn{1}{l}{-}      &  10$\times$5 \\
                                         &             & K600  &       \multicolumn{1}{l}{~~-}       & \textbf{86.1}         &    \multicolumn{1}{l}{-}      &     \multicolumn{1}{l}{-}     &       \multicolumn{1}{l}{-}      &     10$\times$5         \\

ViViT~\cite{arnab2021vivit}              & ImageNet21k &K400   & 81.3              &    \multicolumn{1}{l}{-}       &     \multicolumn{1}{l}{-}     &      \multicolumn{1}{l}{-}    &          \multicolumn{1}{l}{-}   &  4$\times$3  \\
                                         &             &K600   &         \multicolumn{1}{l}{~~-}       & 83.0         &      \multicolumn{1}{l}{-}    &    \multicolumn{1}{l}{-}      &       \multicolumn{1}{l}{-}      &       4$\times$3      \\
                                         &             &SSv2   &         \multicolumn{1}{l}{~~-}       &      \multicolumn{1}{l}{-}     &     \multicolumn{1}{l}{-}     & \textbf{65.9}        &         \multicolumn{1}{l}{-}    &     4$\times$3          \\
                                         &             &MiT    &     \multicolumn{1}{l}{~~-}           &    \multicolumn{1}{l}{-}       &  \multicolumn{1}{l}{-}        &     \multicolumn{1}{l}{-}     &  38.5          &       4$\times$3        \\
VidTr~\cite{zhang2021vidtr}                            & ImageNet21k &K400   & 80.5              &       \multicolumn{1}{l}{-}    &      \multicolumn{1}{l}{-}    &        \multicolumn{1}{l}{-}  &        \multicolumn{1}{l}{-}     &  1$\times$3  \\
                                         &             &K700   &       \multicolumn{1}{l}{~~-}           &     \multicolumn{1}{l}{-}      &  70.8        &      \multicolumn{1}{l}{-}    &    \multicolumn{1}{l}{-}         &      1$\times$3         \\
                                         &             &SSv2   &       \multicolumn{1}{l}{~~-}           &      \multicolumn{1}{l}{-}     &       \multicolumn{1}{l}{-}   & 63.0        &     \multicolumn{1}{l}{-}        &      1$\times$3         \\                                        
TimeSFormer~\cite{bertasius2021space}    & ImageNet21k &K400   & 80.7              &    \multicolumn{1}{l}{-}       &    \multicolumn{1}{l}{-}      &      \multicolumn{1}{l}{-}    &     \multicolumn{1}{l}{-}        &  1$\times$3  \\
                                         &             &K600   &      \multicolumn{1}{l}{~~-}            & 82.2         &  \multicolumn{1}{l}{-}        &      \multicolumn{1}{l}{-}    &   \multicolumn{1}{l}{-}          &       1$\times$3        \\
                                         &             &SSv2   &        \multicolumn{1}{l}{~~-}          &      \multicolumn{1}{l}{-}     &      \multicolumn{1}{l}{-}    & 62.4        &       \multicolumn{1}{l}{-}      &      1$\times$3         \\
\midrule                      
\ourmethod  & ImageNet21k & K400+SSv2+MiT+ImNet &  83.1\textbf{\textsubscript{\textcolor{darkergreen}{(+2.4)}}}              &     \multicolumn{1}{l}{-}          &        \multicolumn{1}{l}{-}      &  64.2       & 41.3           &  1$\times$3  \\
\small{(Based on TimeSFormer)}           &             & K600+SSv2+MiT+ImNet                        &    \multicolumn{1}{l}{~~-}               &     84.5\textbf{\textsubscript{\textcolor{darkergreen}{(+2.3)}}}      &     \multicolumn{1}{l}{-}         &  64.7\textbf{\textsubscript{\textcolor{darkergreen}{(+2.3)}}}       & \textbf{41.5}     &   1$\times$3      \\
           &             & K700+SSv2+MiT+ImNet                        &   \multicolumn{1}{l}{~~-}                &        \multicolumn{1}{l}{-}       &  \textbf{74.9}     &  64.7\textbf{\textsubscript{\textcolor{darkergreen}{(+2.3)}}}       & \textbf{41.5}     &     1$\times$3          \\
\bottomrule
\end{tabular}
\caption{Comparison with the SoTA pretrained on ImageNet-21k. Our results \ourmethod is based on  TimeSFormer architecture. Comparing with TimeSFormer, \ourmethod achieves significant improvements across all datasets. Comparing with ViViT~\cite{arnab2021vivit} and VidTr~\cite{zhang2021vidtr}, \ourmethod improves performance on K400, K600, K700, and MiT datasets. The views are denoted as $\#$ of temporal crops $\times$ $\#$ of spatial crops. Dash (``-'') means the results are not applicable. We note our improvements compared with TimeSFormer of same sizes.}
\label{tab:main_imnet}
\end{table*}

\begin{table*}[t]
    \centering
\begin{tabular}{@{}l@{\quad}l@{}c@{~~}l@{~}l@{~}l@{~}l@{~}l@{~}c@{}}
\toprule
Model                 & Pretrain & Finetune        & K400 & K600 & K700 & SSv2 & MiT & Views \\
\midrule                      
Video SwinTrans~\cite{liu2021video}      & ImageNet21k+K400 & SSv2  &      \multicolumn{1}{l}{-}           &     \multicolumn{1}{l}{-}       &    \multicolumn{1}{l}{-}       & 69.6 &       \multicolumn{1}{l}{-}       &   1$\times$3 \\
TokenLearner~\cite{ryoo2021tokenlearner}          & JFT-300M    & K400  & 85.4     &   \multicolumn{1}{l}{-}         &  \multicolumn{1}{l}{-}         &     \multicolumn{1}{l}{-}      &      \multicolumn{1}{l}{-}        &  4$\times$3  \\
                      &             & K600  &     \multicolumn{1}{l}{-}            & 86.3&   \multicolumn{1}{l}{-}        &     \multicolumn{1}{l}{-}      &      \multicolumn{1}{l}{-}        &     4$\times$3          \\

ViViT~\cite{arnab2021vivit}                 & JFT-300M    &K400   & 84.8              &    \multicolumn{1}{l}{-}        &  \multicolumn{1}{l}{-}         &   \multicolumn{1}{l}{-}        &    \multicolumn{1}{l}{-}          &  4$\times$3  \\
                      &             &K600   &   \multicolumn{1}{l}{-}              & 85.8         &   \multicolumn{1}{l}{-}        &    \multicolumn{1}{l}{-}       &    \multicolumn{1}{l}{-}          &     4$\times$3          \\
MoViNet~\cite{kondratyuk2021movinets} & None & K700 &    \multicolumn{1}{l}{-}            &    \multicolumn{1}{l}{-}        &    72.3     &   \multicolumn{1}{l}{-}        &     \multicolumn{1}{l}{-}         &  1$\times$1  \\
VATT~\cite{akbari2021vatt}              & AudioSet+Videos &K400   & 82.1              &   \multicolumn{1}{l}{-}         & \multicolumn{1}{l}{-}          &   \multicolumn{1}{l}{-}        &    \multicolumn{1}{l}{-}          &  4$\times$3  \\
                      &             &K600   &    \multicolumn{1}{l}{-}             & 83.6         &   \multicolumn{1}{l}{-}        &    \multicolumn{1}{l}{-}       &     \multicolumn{1}{l}{-}         &        4$\times$3       \\
                      &             &MiT    &   \multicolumn{1}{l}{-}              &     \multicolumn{1}{l}{-}       &   \multicolumn{1}{l}{-}        &  \multicolumn{1}{l}{-}         & 41.1  &       4$\times$3        \\
OmniSource~\cite{duan2020omni}         & IG-Kinetics-65M & K400 & 83.6              &     \multicolumn{1}{l}{-}       &   \multicolumn{1}{l}{-}        &     \multicolumn{1}{l}{-}      &      \multicolumn{1}{l}{-}        &  10$\times$3 \\
\midrule                      
\ourmethod & JFT-300M & K400+SSv2+MiT+ImNet & \textbf{86.3}\textbf{\textsubscript{\textcolor{darkergreen}{(+0.9)}}}     &        \multicolumn{1}{l}{-}       &    \multicolumn{1}{l}{-}          &69.3         & \textbf{45.0}\textbf{\textsubscript{\textcolor{darkergreen}{(+3.9)}}}  &  1$\times$3  \\
\small{(Based on TimeSFormer)}           &          & K600+SSv2+MiT+ImNet &   \multicolumn{1}{l}{-}                 &\textbf{86.8}\textbf{\textsubscript{\textcolor{darkergreen}{(+0.5)}}} &   \multicolumn{1}{l}{-}           &\textbf{69.8}\textbf{\textsubscript{\textcolor{darkergreen}{(+0.2)}}}&  44.5          &    1$\times$3     \\
           &          & K700+SSv2+MiT+ImNet &       \multicolumn{1}{l}{-}             &     \multicolumn{1}{l}{-}          & \textbf{78.5}\textbf{\textsubscript{\textcolor{darkergreen}{(+6.2)}}}            &  69.7         &  44.8             &    1$\times$3           \\
\midrule
\ourmethod & JFT-3B   & K400+SSv2+MiT+ImNet & \textbf{87.2}\textbf{\textsubscript{\textcolor{darkergreen}{(+1.8)}}}    &          \multicolumn{1}{l}{-}     &       \multicolumn{1}{l}{-}       & 70.8        & \textbf{46.1}\textbf{\textsubscript{\textcolor{darkergreen}{(+5.0)}}}  &  1$\times$3  \\
\small{(Based on TimeSFormer)}           &          & K600+SSv2+MiT+ImNet &  \multicolumn{1}{l}{-}                  &\textbf{87.9}\textbf{\textsubscript{\textcolor{darkergreen}{(+1.6)}}} &      \multicolumn{1}{l}{-}        &\textbf{70.9}\textbf{\textsubscript{\textcolor{darkergreen}{(+1.3)}}}&  45.9          &     1$\times$3          \\
           &          & K700+SSv2+MiT+ImNet & \multicolumn{1}{l}{-}  & \multicolumn{1}{l}{-}  & \textbf{79.8}\textbf{\textsubscript{\textcolor{darkergreen}{(+7.5)}}} & 70.6 & 45.9                &     1$\times$3          \\
\bottomrule
\end{tabular}
\caption{Comparison with the SoTA pretrained on larger-scale datasets. \ourmethod pretrained on JFT-300M surpassed all SoTA by a margin. \ourmethod pretrained on even larger dataset (JFT-3B) established a new set of SoTA for all datasets. Dash (``-'') means the results are not applicable. We note our improvements compared with the previous state-of-the-arts.}
\label{tab:main_larger}
\vspace{-10pt}
\end{table*}
We summarize the \ourmethod performance in Table~\ref{tab:main_imnet} and Table~\ref{tab:main_larger}. We reported our model performance under three pre-training datasets, ImageNet-21k, JFT-300M, and JFT-3B. For each pre-training setting, we co-train our model on SSv2, MiT, ImageNet, and different versions of Kinetics jointly. 

We compare \ourmethod with the TimeSFormer under the same ImageNet-21k pre-training setting in Table~\ref{tab:main_imnet}. With the same architecture, \ourmethod co-trained on multiple datasets achieves 2.4\%, 2.3\%, and 2.3\% improvement on K400, K600, and SSv2, respectively. Compared with ViViT and VidTr, \ourmethod improves K400, K600, K700, and MiT performance by a margin, which verifies the effectiveness of the proposed approach. \ourmethod achieves lower performance than Video SwinTrans, due to its more advanced architecture. 

When pretrained on a larger-scale image dataset, JFT-300M, \ourmethod surpassed the previous best performance by 0.9\%, 0.5\%, 0.2\%, and 3.9\% on K400, K600, SSv2, and MiT in top-1 accuracy. Pretraining on an even larger image dataset, JFT-3B, further boosts the top-1 accuracy to 87.2\% on K400, 87.9\% on K600, 79.8\% on K700, 70.9\% on SSv2, and 46.1\% on MiT. Our results indicate that training methodology for the transformer model is important. Orthogonal to improving model architecture, co-training a simple spatio-temporal transformer with multiple datasets could achieve superior performance. 

\ourmethod didn't alter the architecture or evaluation scheme. Thus, \ourmethod has the same inference speed as TimeSFormer.

\subsection{Ablation study}

\ourmethod improves the SoTA on Top-1 accuracy across all datasets. In this section, we conduct ablation studies on the \ourmethod pretrained on JFT-3B to empirically analyze the performance gain of \ourmethod. 

\paragraph{Co-training with multiple datasets} 
\begin{table}[]
\centering
\begin{tabular}{@{}l@{\quad\quad\quad\quad\quad}lll@{}}

\multicolumn{4}{c}{\textbf{(a)} The ablation of co-training on K400+SSv2+MiT+ImNet} \\

\toprule
Training strategy      & K400  & SSv2  & MiT   \\
\midrule
Train independently    & 85.0 & 67.2  & 44.9 \\
K400+SSv2              & 86.4 & 69.9  & -    \\
K400+SSv2+MiT          & \textbf{87.2} & 70.6  & 45.9 \\
K400+SSv2+MiT+ImNet    & \textbf{87.2} & \textbf{70.8}  & \textbf{46.1} \\
\bottomrule

\addlinespace
\addlinespace

\multicolumn{4}{c}{\textbf{(b)} The ablation of co-training on K600+SSv2+MiT+ImNet} \\
\toprule
Training strategy      & K600  & SSv2  & MiT   \\
\midrule
Train independently    & 86.0 & 67.2 & 44.9 \\
K600+SSv2              & 87.3 & 70.4 & -    \\
K600+SSv2+MiT          & \textbf{87.9} & 70.5 & \textbf{45.9} \\
K600+SSv2+MiT+ImNet    & \textbf{87.9} & \textbf{70.9} & \textbf{45.9} \\
\bottomrule

\addlinespace
\addlinespace

\multicolumn{4}{c}{\textbf{(c)} The ablation of co-training on K700+SSv2+MiT+ImNet} \\
\toprule
Training strategy      & K700  & SSv2  & MiT   \\
\midrule
Train independently    & 78.0 & 67.2 & 44.9 \\
K700+SSv2              & 79.1 & 70.6 & -    \\
K700+SSv2+MiT          & 79.6 & \textbf{70.7} & 45.8 \\
K700+SSv2+MiT+ImNet    & \textbf{79.8} & 70.6 & \textbf{45.9} \\
\bottomrule

\end{tabular}
\caption{Detailed co-training Top-1 accuracy with model pretrained on JFT-3B. By co-training with more datasets, \ourmethod keeps improve performance on K400, K600, K700, SSv2, and MiT datasets. }
\label{tab:cotrain}
\end{table}
\begin{table}[]
\centering

\begin{tabular}{@{}ll@{\quad}llll@{}}
\multicolumn{6}{c}{Models finetuned on K400+SSv2+MiT+ImNet} \\
\toprule
$w_\texttt{image}$ & $w_\texttt{video}$ & K400  & SSv2  & MiT & ImageNet\\
\midrule
0 & 1    & \textbf{87.2} & 70.6  & 45.9 & - \\
0.5 & 1  & \textbf{87.2} & \textbf{70.8}  & \textbf{46.1} & 86.1\\
0.75 & 1 & \textbf{87.2} & 70.6  & 45.8 & \textbf{86.6} \\
\bottomrule
\end{tabular}
\caption{Top-1 accuracy of \ourmethod under different image classification loss weights. With larger $w_\texttt{image}$, the model improves the ImageNet performance however sacrifices video dataset results.}
\label{tab:lossweight}
\vspace{-10pt}
\end{table}
\vspace{-10pt}

The main argument is that co-training over multiple image and video datasets can improve action recognition performance. To verify this argument, we summarize the ablation results in Table~\ref{tab:cotrain}. 

We first limit the scope to two datasets, K400 and SSv2. We compare the co-training performance with the results of training independently. Co-training improves the top-1 accuracy on K400 by 1.4\% and SSv2 by 2.7\%, which indicates that jointly learning K400 and SSv2 could  enhance the performance of both tasks. We further include the MiT dataset into co-training. Compared with co-training on just two datasets, co-training on all three improves K400 and SSv2 performance by 0.8\% and 0.7\%, respectively. We observe that co-training with MiT also improves MiT performance by 1.0\%. Finally, we co-train both image and video datasets together. Adding ImageNet in the co-training datasets further improves SSv2 and MiT by 0.2\% and 0.2\%, establishing a new SoTA. We observe a similar improvement on the model co-trained with K600+SSv2+MiT+ImageNet and K700+SSv2+MiT+ImageNet. 

\vspace{-10pt}
\paragraph{Loss weight for image and video classification}
Our co-training paradigm involves two losses, the image classification loss and the video classification loss. We study the performance under different loss weights in Table \ref{tab:lossweight}. By increasing the image classification loss weight, we observe that the model is encouraged to learn better appearance representations. Thus, ImageNet accuracy improves from 86.1\% to 86.6\%. However, as the model focuses on learning better appearance information, the model's ability to capture motion structure is reduced. We observed 0.2\% top-1 accuracy drops on SSv2 and MiT datasets, which indicates it is vital to balance the image and video classification loss weight. 

\vspace{-10pt}
\paragraph{Transfer learning on other datasets}

\begin{figure}[t]
\begin{subfigure}{0.5\textwidth}
  \centering
  \includegraphics[width=.9\linewidth]{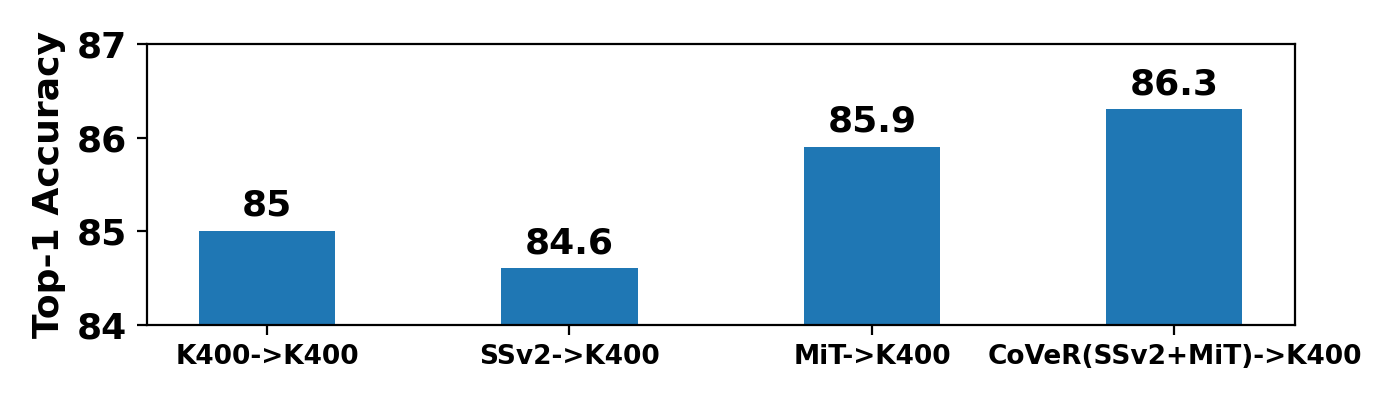}  
  \caption{Transfer learning on K400}
\end{subfigure}
\begin{subfigure}{0.5\textwidth}
  \centering
  \includegraphics[width=0.9\linewidth]{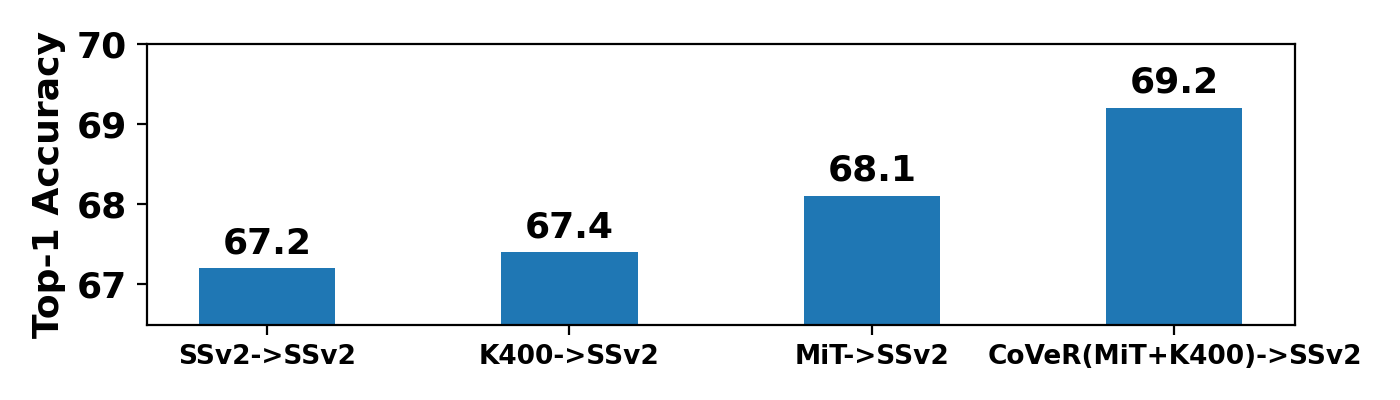}  
  \caption{Transfer learning on SSv2}
\end{subfigure}
\caption{Comparison of transfer learning the representation learned by \ourmethod and standard training paradigm. A$\rightarrow$B means the model is trained on dataset A and then finetuned on dataset B. }
\label{fig:transfer}
\vspace{-10pt}
\end{figure}

We use transfer learning as a showcase to verify the video representation quality. Specifically, we trained on the source datasets, then finetuned and evaluated on the target dataset. The results are summarized in Fig.~\ref{fig:transfer}. We first consider K400 as the target dataset. \ourmethod co-trained on SSv2 and MiT improves the top-1 accuracy on K400$\rightarrow$K400 by 1.3\%, SSv2$\rightarrow$K400 by 1.7\%, and MiT$\rightarrow$K400 by 0.4\%. Similarly, we observe that by transferring to SSv2, \ourmethod achieves 2\%, 1.8\%, and 1.1\% improvement over SSv2$\rightarrow$SSv2, K400$\rightarrow$SSv2, and MiT$\rightarrow$SSv2, respectively. Improvement on transfer learning shows that \ourmethod co-trained on multiple datasets could learn better visual representations than the standard training paradigm, which is useful for downstream tasks.

\vspace{-5pt}
\section{Conclusion}
\label{sec:conclusion}

 In this work, we present \ourmethod, a training paradigm that jointly learns action recognition and object recognition tasks in a single model for the purpose of constructing a general-purpose action recognition framework. Our analysis indicates it may be beneficial to integrate many video datasets into one multi-task learning paradigm. We highlight the importance of continuing to learn on image data during fine-tuning to maintain robust spatial representations. Our empirical findings suggest \ourmethod can learn a single model which achieves impressive performance across many action recognition datasets without an additional stage of fine-tuning on each downstream application. 
 In particular, \ourmethod sets a new state-of-the-art performance on Kinetics, SSv2 and MiT.

{\small
\bibliographystyle{ieee_fullname}
\bibliography{egbib}

\begin{thebibliography}{10}\itemsep=-1pt

\bibitem{akbari2021vatt}
Hassan Akbari, Linagzhe Yuan, Rui Qian, Wei-Hong Chuang, Shih-Fu Chang, Yin
  Cui, and Boqing Gong.
\newblock Vatt: Transformers for multimodal self-supervised learning from raw
  video, audio and text.
\newblock {\em arXiv preprint arXiv:2104.11178}, 2021.

\bibitem{arnab2021vivit}
Anurag Arnab, Mostafa Dehghani, Georg Heigold, Chen Sun, Mario Lu{\v{c}}i{\'c},
  and Cordelia Schmid.
\newblock Vivit: A video vision transformer.
\newblock {\em ICCV}, 2021.

\bibitem{bertasius2021space}
Gedas Bertasius, Heng Wang, and Lorenzo Torresani.
\newblock Is space-time attention all you need for video understanding?
\newblock {\em NeurIPS}, 2021.

\bibitem{carreira2018short}
Joao Carreira, Eric Noland, Andras Banki-Horvath, Chloe Hillier, and Andrew
  Zisserman.
\newblock A short note about kinetics-600.
\newblock {\em arXiv preprint arXiv:1808.01340}, 2018.

\bibitem{carreira2019short}
Joao Carreira, Eric Noland, Chloe Hillier, and Andrew Zisserman.
\newblock A short note on the kinetics-700 human action dataset.
\newblock {\em arXiv preprint arXiv:1907.06987}, 2019.

\bibitem{carreira2017quo}
Joao Carreira and Andrew Zisserman.
\newblock Quo vadis, action recognition? a new model and the kinetics dataset.
\newblock In {\em CVPR}, 2017.

\bibitem{deng2009imagenet}
Jia Deng, Wei Dong, Richard Socher, Li-Jia Li, Kai Li, and Li Fei-Fei.
\newblock Imagenet: A large-scale hierarchical image database.
\newblock In {\em CVPR}, 2009.

\bibitem{devlin2018bert}
Jacob Devlin, Ming-Wei Chang, Kenton Lee, and Kristina Toutanova.
\newblock Bert: Pre-training of deep bidirectional transformers for language
  understanding.
\newblock {\em NAACL}, 2019.

\bibitem{dosovitskiy2020image}
Alexey Dosovitskiy, Lucas Beyer, Alexander Kolesnikov, Dirk Weissenborn,
  Xiaohua Zhai, Thomas Unterthiner, Mostafa Dehghani, Matthias Minderer, Georg
  Heigold, Sylvain Gelly, et~al.
\newblock An image is worth 16x16 words: Transformers for image recognition at
  scale.
\newblock {\em ICLR}, 2021.

\bibitem{duan2020omni}
Haodong Duan, Yue Zhao, Yuanjun Xiong, Wentao Liu, and Dahua Lin.
\newblock Omni-sourced webly-supervised learning for video recognition.
\newblock In {\em ECCV}, 2020.

\bibitem{fan2021multiscale}
Haoqi Fan, Bo Xiong, Karttikeya Mangalam, Yanghao Li, Zhicheng Yan, Jitendra
  Malik, and Christoph Feichtenhofer.
\newblock Multiscale vision transformers.
\newblock {\em arXiv preprint arXiv:2104.11227}, 2021.

\bibitem{ging2020coot}
Simon Ging, Mohammadreza Zolfaghari, Hamed Pirsiavash, and Thomas Brox.
\newblock Coot: Cooperative hierarchical transformer for video-text
  representation learning.
\newblock In {\em NeurIPS}, 2020.

\bibitem{goyal2017something}
Raghav Goyal, Samira Ebrahimi~Kahou, Vincent Michalski, Joanna Materzynska,
  Susanne Westphal, Heuna Kim, Valentin Haenel, Ingo Fruend, Peter Yianilos,
  Moritz Mueller-Freitag, et~al.
\newblock The" something something" video database for learning and evaluating
  visual common sense.
\newblock In {\em ICCV}, 2017.

\bibitem{kay2017kinetics}
Will Kay, Joao Carreira, Karen Simonyan, Brian Zhang, Chloe Hillier, Sudheendra
  Vijayanarasimhan, Fabio Viola, Tim Green, Trevor Back, Paul Natsev, et~al.
\newblock The kinetics human action video dataset.
\newblock {\em arXiv preprint arXiv:1705.06950}, 2017.

\bibitem{kolesnikov2020big}
Alexander Kolesnikov, Lucas Beyer, Xiaohua Zhai, Joan Puigcerver, Jessica Yung,
  Sylvain Gelly, and Neil Houlsby.
\newblock Big transfer (bit): General visual representation learning.
\newblock In {\em ECCV}, 2020.

\bibitem{kondratyuk2021movinets}
Dan Kondratyuk, Liangzhe Yuan, Yandong Li, Li Zhang, Mingxing Tan, Matthew
  Brown, and Boqing Gong.
\newblock Movinets: Mobile video networks for efficient video recognition.
\newblock In {\em CVPR}, 2021.

\bibitem{krishna2017dense}
Ranjay Krishna, Kenji Hata, Frederic Ren, Li Fei-Fei, and Juan Carlos~Niebles.
\newblock Dense-captioning events in videos.
\newblock In {\em CVPR}, 2017.

\bibitem{li2021value}
Linjie Li, Jie Lei, Zhe Gan, Licheng Yu, Yen-Chun Chen, Rohit Pillai, Yu Cheng,
  Luowei Zhou, Xin~Eric Wang, William~Yang Wang, et~al.
\newblock Value: A multi-task benchmark for video-and-language understanding
  evaluation.
\newblock {\em arXiv preprint arXiv:2106.04632}, 2021.

\bibitem{liu2021swin}
Ze Liu, Yutong Lin, Yue Cao, Han Hu, Yixuan Wei, Zheng Zhang, Stephen Lin, and
  Baining Guo.
\newblock Swin transformer: Hierarchical vision transformer using shifted
  windows.
\newblock {\em ICCV}, 2021.

\bibitem{liu2021video}
Ze Liu, Jia Ning, Yue Cao, Yixuan Wei, Zheng Zhang, Stephen Lin, and Han Hu.
\newblock Video swin transformer.
\newblock {\em arXiv preprint arXiv:2106.13230}, 2021.

\bibitem{lu2019vilbert}
Jiasen Lu, Dhruv Batra, Devi Parikh, and Stefan Lee.
\newblock Vilbert: Pretraining task-agnostic visiolinguistic representations
  for vision-and-language tasks.
\newblock {\em NeurIPS}, 2019.

\bibitem{monfort2019moments}
Mathew Monfort, Alex Andonian, Bolei Zhou, Kandan Ramakrishnan, Sarah~Adel
  Bargal, Tom Yan, Lisa Brown, Quanfu Fan, Dan Gutfreund, Carl Vondrick, et~al.
\newblock Moments in time dataset: one million videos for event understanding.
\newblock {\em PAMI}, 2019.

\bibitem{ryoo2021tokenlearner}
Michael~S Ryoo, AJ Piergiovanni, Anurag Arnab, Mostafa Dehghani, and Anelia
  Angelova.
\newblock Tokenlearner: What can 8 learned tokens do for images and videos?
\newblock {\em NeurIPS}, 2021.

\bibitem{schuldt2004recognizing}
Christian Schuldt, Ivan Laptev, and Barbara Caputo.
\newblock Recognizing human actions: a local svm approach.
\newblock In {\em ICPR}, 2004.

\bibitem{simonyan2014two}
Karen Simonyan and Andrew Zisserman.
\newblock Two-stream convolutional networks for action recognition in videos.
\newblock {\em NeurIPS}, 2014.

\bibitem{su2019vl}
Weijie Su, Xizhou Zhu, Yue Cao, Bin Li, Lewei Lu, Furu Wei, and Jifeng Dai.
\newblock Vl-bert: Pre-training of generic visual-linguistic representations.
\newblock {\em ICLR}, 2020.

\bibitem{touvron2021training}
Hugo Touvron, Matthieu Cord, Matthijs Douze, Francisco Massa, Alexandre
  Sablayrolles, and Herv{\'e} J{\'e}gou.
\newblock Training data-efficient image transformers \& distillation through
  attention.
\newblock In {\em ICML}, 2021.

\bibitem{tran2015learning}
Du Tran, Lubomir Bourdev, Rob Fergus, Lorenzo Torresani, and Manohar Paluri.
\newblock Learning spatiotemporal features with 3d convolutional networks.
\newblock In {\em ICCV}, 2015.

\bibitem{wang2021tdn}
Limin Wang, Zhan Tong, Bin Ji, and Gangshan Wu.
\newblock Tdn: Temporal difference networks for efficient action recognition.
\newblock In {\em CVPR}, 2021.

\bibitem{zhai2021scaling}
Xiaohua Zhai, Alexander Kolesnikov, Neil Houlsby, and Lucas Beyer.
\newblock Scaling vision transformers.
\newblock {\em arXiv preprint arXiv:2106.04560}, 2021.

\bibitem{zhang2021vidtr}
Yanyi Zhang, Xinyu Li, Chunhui Liu, Bing Shuai, Yi Zhu, Biagio Brattoli, Hao
  Chen, Ivan Marsic, and Joseph Tighe.
\newblock Vidtr: Video transformer without convolutions.
\newblock In {\em CVPR}, 2021.

\end{thebibliography}
}

\clearpage
\appendix
\label{sec:supp}
In the supplementary material, we provide details and experiments omitted from the main text due to the limited space, including:
\begin{enumerate}
    \item \S~\ref{supp:sec:details} describes the implementation details for the co-training on video and image data (\S 3.3.2 in the main text).
    \item \S~\ref{supp:sec:exp} provides additional ablation studies on \ourmethod. Specifically, we study the performance under different input configurations and different model variants.
\end{enumerate}

\section{Implementation Details of Co-train Video and Image Data}
\label{supp:sec:details}
We provide the details of implementing the co-training on multiple video and image datasets.
Without modifying the architecture, we calculate the image and video classification losses by feed-forwarding the images and videos separately. 

We consider the images as single-frame videos. To calculate the image classification loss, we take images as input and obtain the loss $\ell(\{y^j_{\texttt{image}}\},  \{c^j_\texttt{image}\})$. As images are considered as single-frame videos, the key set K, the query set Q, and the value set V in the temporal multi-head attention layer are the same. Temporal multi-head attention will be a feed-forward network. 

After obtaining the image loss, we feed-forward the video to the model $f$ and obtain the video classification loss $\ell(\{y^j_\texttt{video}\},  \{c^j_\texttt{video}\})$. The co-training loss for video and image data $\ell_\texttt{image\_video}$ is the weighted average of both image and video classification loss. We calculate the gradients and update the weights based on $\ell_\texttt{image\_video}$.

\section{Additional Experiments on \ourmethod}
\label{supp:sec:exp}

In this section, we reported the ablation studies that are omitted from the main paper due to space limitations. Specifically, we studied \ourmethod performance under different model variants and different input configurations.

\paragraph{Model variants}
\begin{figure}[t]
\includegraphics[width=\linewidth]{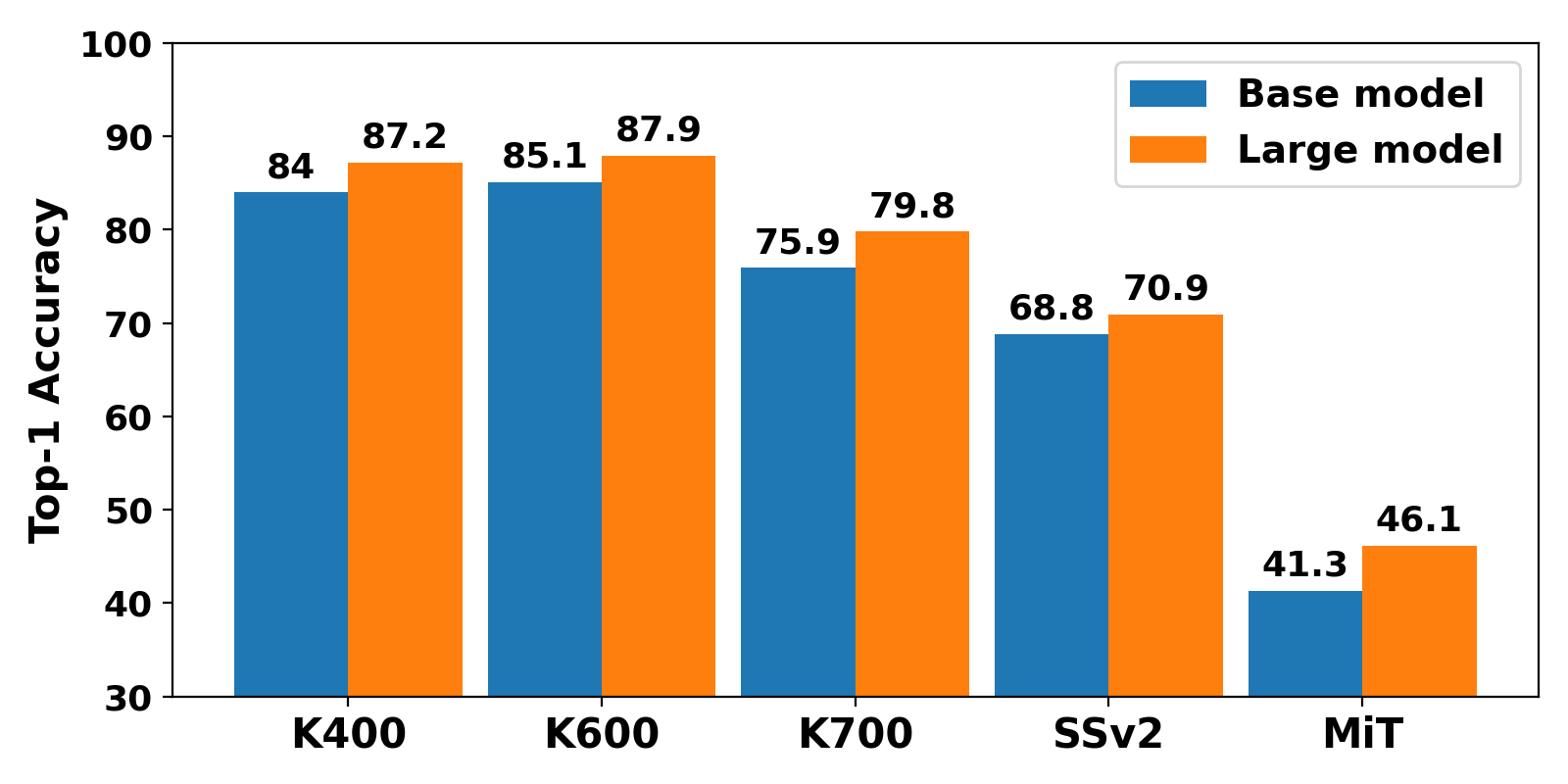}  
\caption{Comparison of base model and large model using \ourmethod pretrained on JFT-3B. The large model constantly improves over the base model on all datasets.}
\label{fig:variants}
\end{figure}
We first compare the performance under the different sizes of architectures. We consider the TimeSFormer-Base model for comparison. The TimeSFormer-Base architecture is similar to the TimeSFormer-Large model, which contains 12 TimeSFormer blocks. Each block includes one temporal multi-head attention layer and one spatial multi-head attention layer. The multi-head attention layer has 12 attention heads with 768 hidden dimensions. We follow the same learning rate scheduling and apply the same data augmentation to train the TimeSFormer-Base model as described in Sect. 4.1 in the main paper.

We summarize the comparison between TimeSFormer-Large and TimeSFormer-Base model in Fig.~\ref{fig:variants}. We observe that with co-training, the large model improves the top-1 accuracy by $3.2\%$, $2.8\%$, $3.9\%$, $2.1\%$, $5.2\%$ on K400, K600, K700, SSv2, and MiT. By enlarging the model capacity, we didn't experience overfitting. Here, we empirically show that \ourmethod could unleash the model performance by leveraging a large amount of training data. With the co-training, the large model shows solid improvement over the base model by a large margin across all datasets.

\paragraph{Input configuration}
\begin{table}[t]
\centering
\begin{tabular}{@{}l@{\quad\quad\quad\quad\quad\quad}lll@{}}

\multicolumn{4}{c}{\textbf{(a)} Models trained on K400+SSv2+MiT+ImNet} \\

\toprule
Input setting   & K400  & SSv2  & MiT   \\
\midrule
Standard        & 85.1 & 67.4 & 44.2 \\
High Resolution & \textbf{87.2} & \textbf{70.8} & \textbf{46.1} \\
\bottomrule

\addlinespace
\addlinespace
\multicolumn{4}{c}{\textbf{(b)} Models trained on K600+SSv2+MiT+ImNet} \\

\toprule
Input setting   & K600  & SSv2  & MiT   \\
\midrule
Standard        & 85.6 & 67.7 & 44.0 \\
High Resolution & \textbf{87.9} & \textbf{70.9} & \textbf{45.9} \\
\bottomrule

\addlinespace
\addlinespace
\multicolumn{4}{c}{\textbf{(b)} Models trained on K700+SSv2+MiT+ImNet} \\

\toprule
Input setting   & K700  & SSv2  & MiT   \\
\midrule
Standard        & 76.2 & 67.9 & 43.9 \\
High Resolution & \textbf{79.8} & \textbf{70.6} & \textbf{45.9} \\
\bottomrule
\end{tabular}
\caption{Comparison of different input settings for co-training on Kinetics, SSv2, and MiT datasets.}
\label{tab:input}
\vspace{-10pt}
\end{table}
Next, we analyze the model performance under different input configurations. Here, we follow TimeSFormer\cite{bertasius2021space} to consider two input setting, standard and high resolution. We sample $8$ frames from the video for the standard setting, and the input patch resolution is $224\times224$. For the high-resolution setting, $16$ frames are sampled from video. The resolution for the input patch is $448 \times 448$. We summarize the results of the model pre-trained on JFT-3B in Table \ref{tab:input}. The model with high resolution constantly improves the performance by a large margin across all benchmarks.  

\end{document}